\documentclass{article}
\usepackage{amsmath}
\usepackage{booktabs}
\usepackage{multirow}
\usepackage{graphicx}
\usepackage{wrapfig}
\usepackage{enumitem}
\usepackage{caption}
\usepackage{amssymb}
\usepackage{tabularx}
\usepackage{booktabs}
\usepackage{makecell}
\usepackage{multirow}
\usepackage{float}
\usepackage{xcolor}
\usepackage{hyperref}
\definecolor{lightblue}{RGB}{0,120,200}
\usepackage[preprint]{corl_2026}

\title{ZeroWBC: Learning Natural Whole-Body Humanoid Interaction from Human Egocentric Data}
\author{
Haoran Yang\textsuperscript{1,2,*} \quad
Jiacheng Bao\textsuperscript{2,*} \quad
Yucheng Xin\textsuperscript{2,3,*} \\
\textbf{Haoming Song\textsuperscript{2,4} \quad
Yuyang Tian\textsuperscript{1,2} \quad
Bin Zhao\textsuperscript{2} \quad
Dong Wang\textsuperscript{2,\textdagger} \quad
Xuelong Li\textsuperscript{5}}
\\[0.6em]
\textsuperscript{1}University of Science and Technology of China \quad
\textsuperscript{2}Shanghai AI Laboratory \\
\textsuperscript{3}Tsinghua University \quad
\textsuperscript{4}Shanghai Jiao Tong University \quad
\textsuperscript{5}TeleAI, China Telecom
\\[0.5em]
\textsuperscript{*}Equal contribution \qquad
\textsuperscript{\textdagger}Corresponding author
\\[0.5em]
Project Website: \href{https://ZeroWBC.github.io}{\textcolor{lightblue}{ZeroWBC.github.io}}
}

\begin{document}
\maketitle

\vspace{-2.0em}
\begin{center}
    \includegraphics[width=\textwidth]{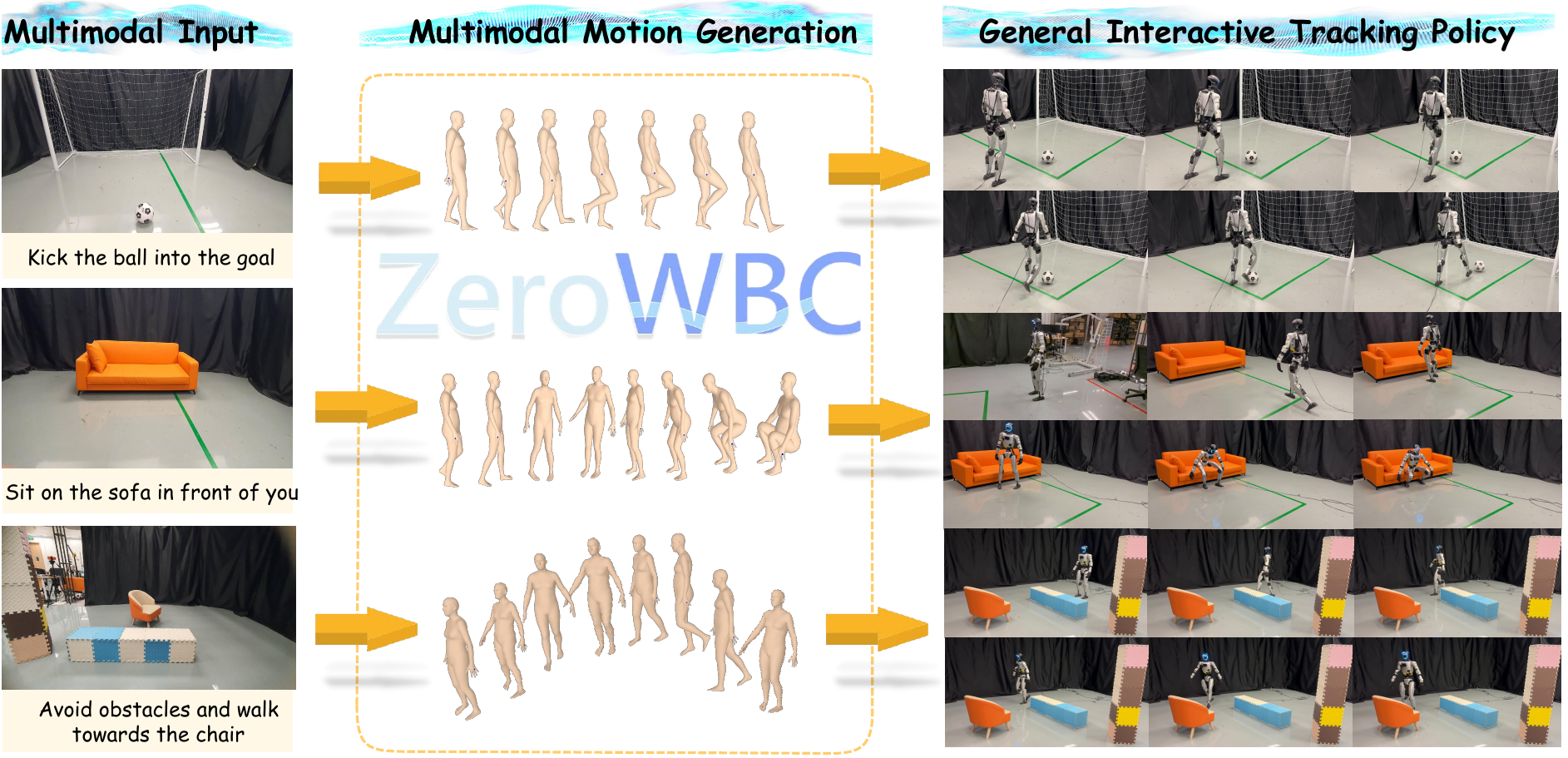}
    \captionof{figure}{
    Overview of the ZeroWBC framework. The pipeline takes an initial egocentric image and text instruction as input (Left), synthesizes human whole-body motions via a fine-tuned vision-language model (Middle), and executes robot actions on the Unitree G1 robot using a general interactive tracking policy (Right).
    }
    \label{fig:overview}
\end{center}

\vspace{-0.8em}


\begin{abstract}
Achieving versatile and natural whole-body humanoid interaction control remains challenging due to the high cost of whole-body teleoperation data. We present ZeroWBC, a teleoperation-free framework that learns humanoid whole-body interaction from human egocentric videos paired with synchronized whole-body motion and text annotations. ZeroWBC adopts a generation-then-tracking formulation to tackle the static scene whole-body interaction control problem. Given an initial egocentric image and a language instruction, a fine-tuned Vision-Language Model generates future human whole-body motion tokens, which are decoded into continuous motions and retargeted to the humanoid. The resulting reference motions, together with root and key body-part trajectories, are then executed by a general interactive motion tracking policy. To improve interaction performance, we introduce an interaction-oriented tracking reward that prioritizes global root and key body-part trajectory alignment while preserving natural whole-body motion. Experiments on the Unitree G1 humanoid robot show that ZeroWBC enables diverse scene-aware behaviors without robot teleoperation demonstrations. These results suggest a scalable paradigm for learning natural humanoid whole-body interaction from human egocentric data.
\end{abstract}

\keywords{Whole-body Control, Motion Generation, General Motion Tracking}


\section{Introduction}
\label{sec:intro}
Humanoid robots are expected to perform scene-aware whole-body interactions in complex real-world environments. Recent progress in humanoid control has improved the fidelity and robustness of tracking reference motions~\cite{liao2025beyondmimic,xie2025kungfubot,he2025asap,huang2025adaptable,margolis2025softmimic,zeng2025behaviora}, while large-scale motion generation and generation-to-control pipelines~\cite{shao2025langwbc,li2025language,yue2025rl,li2026w1,jiang2025uniact,luo2025sonic} have enabled diverse humanoid behaviors. However, most existing systems remain motion-centric, focusing on locomotion, dancing, or athletic skills, and thus provide limited capability for interacting with surrounding objects and scene geometry. Existing visual humanoid control systems typically rely on either large-scale robot teleoperation data~\cite{fu2024humanplus,jiang2025wholebodyvla,ze2025twist2} or task-specific reinforcement learning in simulation~\cite{yin2025visualmimic,xue2025opening}. The former is expensive to collect and often uses decoupled upper and lower-body control~\cite{jiang2025wholebodyvla,ben2025homie,cheng2024expressive,li2025amo}, limiting natural coordination, while the latter is restricted to narrow tasks and vulnerable to sim-to-real gaps. 

The high cost of whole-body teleoperation data motivates human demonstrations as an alternative supervision source. Human egocentric videos paired with synchronized whole-body motions and text annotations provide first-person scene observations, dense whole-body motion labels, and task-level semantics at lower cost. In real-world setup, humanoid robot whole-body teleoperation typically requires two operators to collect about 100 demonstrations in 8 hours, whereas a single human demonstrator can collect about 300 egocentric motion demonstrations in 2 hours in the same task. 

To transform human demonstrations into executable humanoid behaviors, we adopt a generation-then-tracking formulation. A high-level model generates all future human whole-body interaction motions from an initial egocentric image and a language instruction, while a low-level tracking policy retargets and executes all motions at once. This open-loop design is practical for current Vision-Language models, whose inference latency makes high-frequency closed-loop control difficult, and is well suited to many mostly static interaction tasks where the initial observation provides sufficient spatial context. Moreover, preserving interaction precision after human-to-humanoid transfer remains challenging due to morphology, joint-limit, and contact differences. Pure joint-space tracking may reproduce retargeted poses but misalign task-critical points, such as hands for object transport, or root for sitting. We therefore train a general interactive motion tracking policy that tracks both the retargeted robot motion and the human-derived global root and key body-part trajectories, preserving task-critical spatial intent while maintaining natural whole-body motion.

To this end, we introduce \textbf{ZeroWBC}, a two-stage framework for teleoperation-free humanoid static scene whole-body interaction control. As illustrated in Figure~\ref{fig:overview}, ZeroWBC first discretizes human motions into motion tokens using a VQ-VAE tokenizer. A fine-tuned Vision-Language Model then autoregressively predicts all future whole-body motion tokens conditioned on the initial egocentric observation and language instruction. The generated motions are decoded, retargeted to the humanoid, and executed by the general interactive tracking policy using global root and key body-part trajectories with retargeted robot motions. By learning from human egocentric videos with synchronized human whole-body motions and text annotations, ZeroWBC enables diverse static scene-aware whole-body interaction without large-scale robot teleoperation demonstrations.

In summary, our contributions are:
1) \textbf{Teleoperation-free for humanoid interaction:} 
We introduce a scalable supervision paradigm that learns scene-aware humanoid whole-body interaction from human egocentric videos paired with synchronized human motion and text annotations, avoiding the need for costly robot teleoperation demonstrations. 
2) \textbf{Generation-to-Tracking framework:} 
We propose ZeroWBC, a two-stage framework that converts an initial egocentric image and a language instruction into executable humanoid whole body interaction behaviors.
3) \textbf{Interaction-oriented general tracking :} 
We design a tracking objective that emphasizes global root and key body-part trajectory alignment, enabling the humanoid to better preserve task-critical interaction intent during physical execution.

\section{Related Work}
\label{sec:related_work}
\noindent\textbf{Multimodal Motion Generation.}
Text-conditioned human motion generation has become increasingly mature~\cite{ahn2018text2action,guo2022tm2t,jiang2023motiongpt,zhang2024motiongpt,song2025hume,zhong2023attt2m,zhang2023generating,liang2024omg}. Leveraging the pre-trained knowledge of Large Language Models (LLMs), autoregressive methods such as MotionGPT~\cite{zhang2024motiongpt,jiang2023motiongpt} and T2M-GPT~\cite{zhang2023generating} generate human motions by predicting discrete motion tokens, showing stronger semantic understanding, generalization, and inference efficiency than diffusion-based approaches~\cite{tevet2022human,zhang2024motiondiffuse,chen2023executing}. Recent multimodal frameworks, such as M3GPT~\cite{luo2024m} and MotionCraft~\cite{bian2025motioncraft}, further incorporate music and spatial trajectories as conditions. However, motion generation from egocentric image and language remains largely underexplored due to data scarcity and the lack of suitable deployment platforms. Since real-world humanoid robots rely on egocentric perception, it is crucial to generate human interactive motions from both visual observations and textual instructions.

\noindent\textbf{Humanoid Whole-body Control.}
Humanoid whole-body control has evolved from task-specific reinforcement learning for isolated skills such as robust locomotion~\cite{chen2025learning} and parkour~\cite{zhuang2024humanoid}, toward data-driven motion imitation for tracking human references~\cite{xie2025kungfubot,fu2024humanplus,ji2024exbody2,he2024omnih2o}. Early works such as ExBody~\cite{ji2024exbody2} explored decoupled control, while OmniH2O~\cite{he2024omnih2o} established a teacher-student distillation pipeline for unified whole-body control. More recently, BeyondMimic~\cite{liao2025beyondmimic} achieved high-fidelity tracking of single reference motions. The field is now moving toward General Motion Tracking, which aims to track arbitrary motion references without per-motion reward engineering. Along this line, GMT~\cite{chen2025gmt}, PhyGile~\cite{bao2026phygile}, and SONIC~\cite{luo2025sonic} improve adaptability and scalability across diverse skills, while SONIC~\cite{luo2025sonic}, Twist2~\cite{ze2025twist2} applies a general tracker to holistic whole-body teleoperation.

\noindent\textbf{Visual Humanoid Control.}
Visual humanoid control aims to integrate perception and control, but existing methods remain limited in task complexity, motion naturalness, and data scalability. Keypoint-based methods such as HEAD~\cite{chen2025hand} introduce egocentric vision but are mainly restricted to simple navigation, while VideoMimic~\cite{allshire2025visual} enables interaction through a real2sim2real pipeline but lacks first-person perception and still suffers from a sim-to-real gap. High-level planning methods such as LEVER-B~\cite{xue2025leverb} rely on simulation and third-person observations, making them sensitive to domain shifts and instruction variations. Task-specific methods such as VisualMimic~\cite{yin2025visualmimic} and Doorman~\cite{xue2025opening} also suffer from limited generalization and sim-to-real gaps. Meanwhile, Humanoid-VLA~\cite{zhuang2024humanoid} mainly focuses on locomotion, and WholebodyVLA~\cite{jiang2025wholebodyvla} constrains lower-body motion through decoupled strategies, limiting natural whole-body behaviors. Although Twist2~\cite{ze2025twist2} demonstrates holistic control, it depends on costly, hardware-specific teleoperation data.

\begin{figure*}[t]
    \centering
    \includegraphics[width=\textwidth]{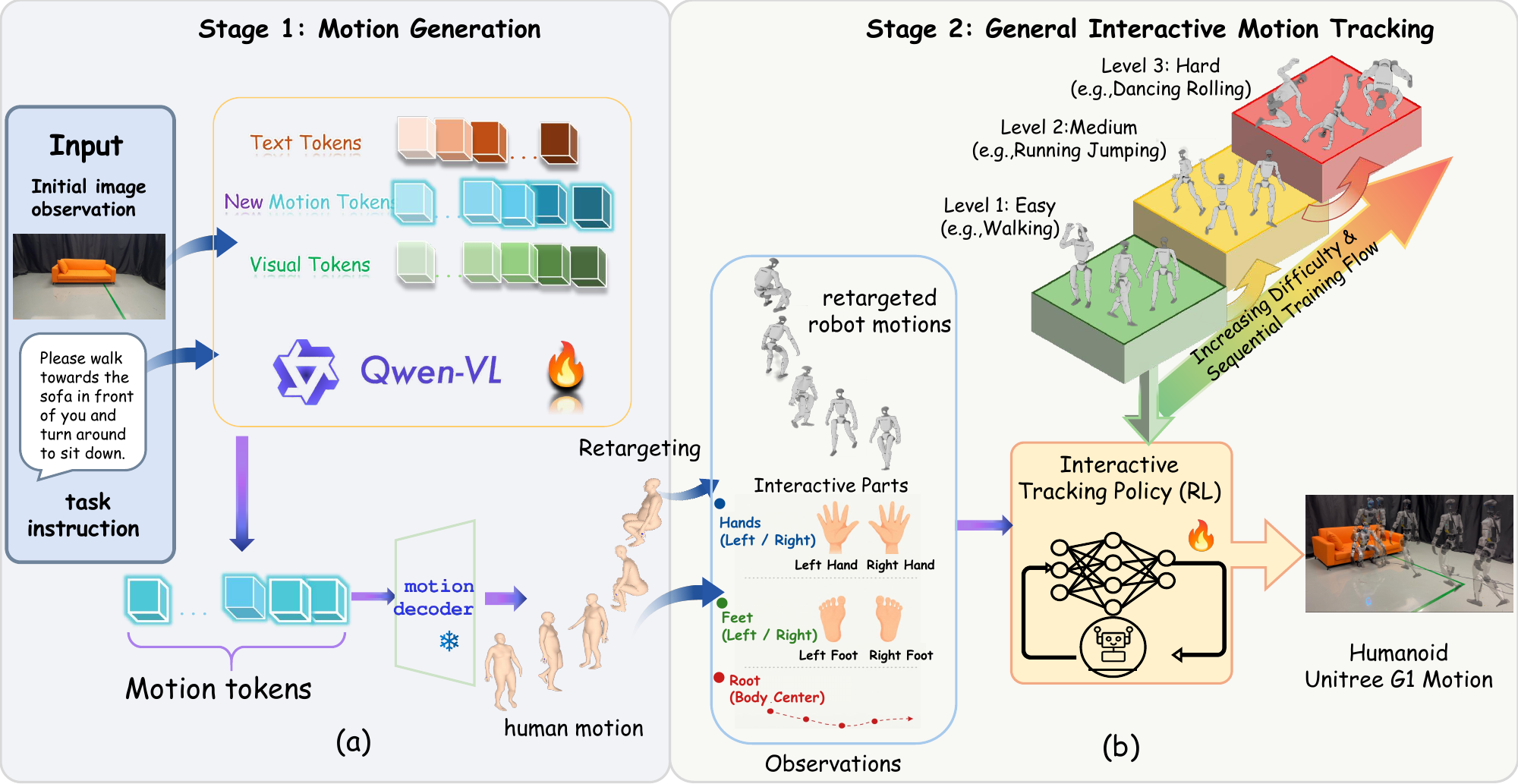}
    \caption{Detailed architecture of ZeroWBC. (a) A fine-tuned Qwen-VL predicts motion tokens from an initial egocentric image and language instruction, which are decoded into continuous human whole-body motions. (b) The generated motions are retargeted to the humanoid and executed by an RL-based policy that tracks both robot reference motions and interaction-relevant body parts.
}
    \label{pipeline}
\end{figure*}


\section{Method}
\label{sec:method}
As illustrated in Fig.~\ref{pipeline}, ZeroWBC adopts a hierarchical two-stage framework for vision-conditioned humanoid whole-body scene-interaction. In the first stage, a VQ-VAE~\cite{vqvae} encodes continuous human motions into discrete motion tokens. A Vision-Language Model is then fine-tuned on image-text-motion pairs for cross-modal alignment, enabling it to predict motion tokens from visual and textual prompts. These tokens are decoded into continuous human motions by a motion decoder. In the second stage, we train a general interactive motion tracker with reinforcement learning and curriculum learning. The tracker maps key interaction position trajectories and robot reference motions derived from human motions into robot joint commands. 

\subsection{Data Collection}
\label{sec:data_collection}
Robot teleoperation data collection is difficult and costly, as it requires precise coordination between the robot and the operator, making large-scale acquisition impractical. We instead adopt a human-centric data collection pipeline. Our setup combines a chest-mounted GoPro camera with an inertial or optical motion capture system, requiring only a MoCap suit and a portable action camera. This design is low-cost and highly scalable. During inference, the robot uses the same camera configuration for perception. To reduce the domain gap between human and robot viewpoints, we align the camera height during data collection with the robot camera height.
\subsection{Motion Tokenizer}
We adopt a VQ-VAE~\cite{vqvae} tailored to SMPL~\cite{SMPL:2015} human-motion data, which compresses continuous high-dimensional motion sequences into compact discrete motion tokens for downstream generation. Given a motion clip of length $n_i$, the encoder downsamples it into a latent sequence of length $n_t$, where $n_i=n_t \cdot 2^{n_{down}}$ and $n_{down}$ is the number of downsampling blocks. The quantized latent tokens are then reconstructed by a lightweight decoder to the original motion resolution. Both the encoder and decoder follow a ResNet-based architecture. For each motion sequence $\mathbf{m}$, we represent its SMPL pose as $\mathbf{m}_{\mathrm{smpl}}\in\mathbb{R}^{66}$ and global translation as $\mathbf{m}_{\mathrm{trans}}\in\mathbb{R}^{3}$. To improve reconstruction quality and physical plausibility, we train the motion VQ-VAE with reconstruction, commitment, velocity, root-rotation, and global-translation losses. 
\subsection{Fine-tuning Vision-Language Model}

We adopt the pre-trained Qwen2.5-VL-3B~\cite{bai2025qwen2} as the backbone for multimodal motion generation. Our goal is to align the discrete motion space with the existing vision-language embedding space of the Qwen-VL, such that whole-body motion generation can be formulated as a standard next-token prediction task. Specifically, we convert the quantized indices from the VQ-VAE codebook into special motion tokens, e.g., $\langle \text{motion\_token\_1} \rangle$, and add them to the Qwen-VL vocabulary.

Given an initial egocentric visual observation $\mathbf{v}$ and a language instruction $\mathbf{t}$, the model autoregressively predicts the complete future motion sequence in one generation pass. This corresponds to our open-loop motion planning setting, where all motion tokens required for the task are generated before execution rather than updated online during tracking. Formally, for a target motion token sequence $\mathbf{z} = \{z_1, z_2, \dots, z_T\}$, we optimize the model parameters $\theta$ by minimizing the negative log-likelihood:
\begin{equation}
\mathcal{L}_{\text{gen}} = - \sum_{i=1}^{T} \log P_\theta(z_i \mid \mathbf{v}, \mathbf{t}, z_{<i}),
\end{equation}
where $z_{<i}$ denotes previously generated motion tokens. The generated token sequence is then decoded by the VQ-VAE decoder into a continuous human motion trajectory.

To improve both generalization and spatial grounding, we use a two-stage fine-tuning strategy. In the first stage, the model is trained on large-scale public image-text-motion data to learn general cross-modal alignment between visual-language inputs and motion tokens. In the second stage, we further fine-tune the model on our self-collected egocentric dataset, which provides higher-quality first-person observations and synchronized whole-body motion labels for interaction tasks. This domain-specific fine-tuning improves the model's ability to generate spatially grounded and physically plausible motions from the robot-aligned egocentric view.
\subsection{General Interactive Motion Tracking}
We train a unified control policy to track diverse whole-body motion capture sequences. To improve stability across heterogeneous motion styles and difficulty levels, we combine adaptive motion scheduling, progressive difficulty exposure, and future motion conditioning.

\noindent\textbf{Tracking objectives.}
We formalize the tracking objective of our motion interactive tracker as a composite target vector:

\[
G=\{\mathbf{g}_{\text{inter}},\mathbf{m}_{\text{ref}}\}\in\mathbb{R}^{D_G}
\]

where $D_G$ denotes the total dimension of the tracking target. The target $G$ is decomposed into two components: the interactive trajectories of key body-parts $\mathbf{g}_{\text{inter}}$ and a retargeted reference motion $\mathbf{m}_{\text{ref}}$.

The interactive trajectories of key body parts encode the global trajectories of key end-effector and humanoid root frames for environment interaction. It is defined as:
\[
\mathbf{g}_{\text{inter}}=\{p_{\text{k}}, \dot p_{k},w_{\text{root}}\}\in\mathbb{R}^{33}
\]

where $p_{\text{k}}\in\mathbb{R}^{3}$, $\dot p_{\text{k}}\in\mathbb{R}^{3}$ denotes the global position and the linear velocity of the corresponding key body part $\text{k}=\{\text{root, left wrist, right wrist, left foot, right foot}\}$, $w_{\text{root}}\in\mathbb{R}^{3}$ means the angular velocity of the root. All these quantities are extracted from the generated human motion sequence in SMPL format.

The reference motion target specifies the joint-level kinematic reference for the robot, obtained via motion retargeting from human to Unitree G1 kinematics. It is given by:

\[
\mathbf{m}_{\text{ref}}=\{q_{\text{ref}},\dot q_{\text{ref}}\}\in\mathbb{R}^{2n_j}
\]

where $q_{\text{ref}}\in\mathbb{R}^{n_j}$ is the reference joint angular vector, $\dot q_{\text{ref}}\in\mathbb{R}^{n_j}$ is the reference joint angular velocity vector, and $n_j=23$ is the number of joints.

The optimization of the tracking objectives is embedded into the design of the imitation learning framework for the general interactive motion tracking policy. As summarized in Table~\ref{tab:obs_space} in Appendix, we augment the observation space with measurements of the interactive trajectories $\mathbf{g}_{\text{inter}}$ of key body parts to explicitly encode task-relevant interaction dynamics. Meanwhile, as detailed in Table~\ref{tab:task_rewards} in Appendix, the reward function is formulated to prioritize the tracking of interactive trajectories while attenuating the penalty for joint-space deviations from the reference motion $\mathbf{m}_{\text{ref}}$. 

This weighted formulation enables the policy to learn the intended motion patterns while enforcing end-effector compliance with the world-space trajectories of critical interaction points. Consequently, the robot maintains precise alignment with task-essential kinematic constraints, leading to improved task completion performance.

\noindent\textbf{Adaptive Motion Scheduling.}
We treat each motion clip as an independent training unit and assign it a dynamic sampling weight according to its current tracking difficulty. For each motion file $i$, we maintain an EMA tracking error $E_i$ and EMA success/failure statistics, from which we estimate the success probability $\hat{p}_i^{\mathrm{succ}}$. The final difficulty score combines persistent tracking error with failure frequency:
\begin{equation}
r_i = (1 - w)\,\mathrm{clip}\!\left(\frac{E_i}{c}, 0, 1\right)
      + w \left(1 - \hat{p}_i^{\mathrm{succ}}\right),
\end{equation}
where $c$ normalizes the tracking error and $w$ balances the error-based and failure-based terms. We convert $r_i$ into sampling probabilities using a temperature-scaled softmax with a uniform exploration term, prioritizing difficult clips while preserving coverage of all motions.

\noindent\textbf{Progressive Difficulty Exposure.}
Because highly complex motions can hinder early-stage convergence, we organize motion files into difficulty levels according to semantic and kinematic complexity and progressively expand the training set. Training starts from the lowest level. Let $l_{\max}$ denote the highest unlocked level. The curriculum advances when the current level satisfies the MPJPE and MPJAE thresholds, or after a fixed number of iterations:
\begin{equation}
\mathrm{MPJPE}_{l_{\max}} < \theta_{\mathrm{pos}}, \quad
\mathrm{MPJAE}_{l_{\max}} < \theta_{\mathrm{ang}}.
\end{equation}
Within the unlocked set, sampling favors the most recently introduced level, while a minimum ratio for each unlocked level mitigates catastrophic forgetting. Newly unlocked clips are introduced gradually to avoid abrupt distribution shifts.

\noindent\textbf{Temporal Conditioning with Future Motion Encoding.}
To anticipate upcoming movements, we augment the policy with a future motion encoder and train it with asymmetric PPO. The critic observes privileged reference and future motion states during training, while the actor uses only compact test-time observations, improving value estimation without increasing deployment complexity. The actor input includes proprioceptive states and a motion command composed of current, short-horizon, and long-horizon target poses. Current and short-horizon targets support near-term tracking, while long-horizon targets are encoded by a lightweight temporal convolutional encoder to capture global motion trends. This multi-scale conditioning helps the policy anticipate velocity changes, directional transitions, and contact events, improving tracking stability for dynamic motions. Additional observation details are provided in the Appendix.

\section{Experiment}
\label{sec:exp}
In this section, we evaluate ZeroWBC on multimodal motion generation, interactive motion tracking, and real-world humanoid interaction tasks on Unitree G1 robot.
\subsection{Dataset}
\label{sec:dataset}
We use three datasets for training and evaluation. \textbf{Nymeria}~\cite{ma2024nymeria} provides 300 hours of synchronized egocentric video, whole-body motion, and language annotations, which we process into 5--10s image-text-SMPL clips for first-stage vision-language-motion alignment. \textbf{HumanML3D}~\cite{Guo_2022_CVPR} contains 14,616 text-annotated 3D motion sequences, and is used to strengthen text-motion alignment and pretrain the interaction tracker after retargeting to Unitree G1 with GMR~\cite{araujo2025retargeting}. \textbf{Self-collected Dataset} contains 5 hours of egocentric data covering navigation, box moving, and sofa sitting under varied scenes and object placements. We use it for second-stage VLM fine-tuning and task-adaptive tracker fine-tuning. More details are provided in the Appendix.

\subsection{Evaluation on Multimodal Motion Generation}
We evaluate the motion generation module on Nymeria and our self-collected dataset using standard metrics, including FID, R-Precision, and MM-Dist. We first train the VQ-VAE on all datasets for motion tokenization, and then fine-tune Qwen2.5-VL-3B in two stages. The first stage uses about 200 hours of image-text-motion data from Nymeria and 25 hours of text-motion data from HumanML3D to learn general visual-language-motion alignment. The second stage further fine-tunes the model on our self-collected egocentric data to improve spatial grounding for interaction tasks. To preserve the pretrained Qwen-VL capability, only the LM decoder is fine-tuned, while the remaining parameters are frozen. This phase utilizes 32 NVIDIA A100 GPUs for training and inference. 
\begin{wraptable}[10]{r}{0.62\textwidth}
\centering
\caption{Motion generation results with different fine-tuning data on Nymeria and self-collected test sets.}
\vspace{-0.6em}
\label{tab:main_comparison}
\small
\setlength{\tabcolsep}{4pt}
\renewcommand{\arraystretch}{1.05}
\begin{tabular}{l l c c c}
\toprule
\textbf{Test Set} & \textbf{Training Data} & \textbf{FID}$\downarrow$ & \textbf{R@3}$\uparrow$ & \textbf{MM-Dist}$\downarrow$ \\
\midrule
\multirow{3}{*}{Nymeria}
& Nymeria              & 0.756 & 0.755 & 3.124 \\
& Nymeria+HumanML3D   & 0.423 & 0.828 & 2.514 \\
& N + H + Self-collected     & \textbf{0.298} & \textbf{0.847} & \textbf{2.286} \\
\midrule
\multirow{3}{*}{Self-Coll}
& Nymeria              & 0.784 & 0.782 & 3.456 \\
& Nymeria+HumanML3D   & 0.628 & 0.825 & 3.128 \\
& N + H + Self-collected      & \textbf{0.245} & \textbf{0.892} & \textbf{2.456} \\
\bottomrule
\end{tabular}
\vspace{-1.0em}
\end{wraptable}
As shown in Table~\ref{tab:main_comparison}, joint training with Nymeria and HumanML3D further improves over single-dataset training, and second-stage fine-tuning on our self-collected data achieves the best performance on both Nymeria and self-collected test sets. These results suggest that large-scale public data provides general motion priors, while self-collected egocentric data improves in-domain interactive spatial grounding. 
\begin{wraptable}[6]{r}{0.54\textwidth}
\centering
\vspace{-2.0ex}
\caption{Evaluation of motion generation quality. We compare ZeroWBC with the baseline MotionGPT.}
\label{tab:motion_metrics}
\vspace{-1.0ex}
\resizebox{0.55\columnwidth}{!}{%
\begin{tabular}{l c c c}
\toprule
\textbf{Method} & \textbf{FID} $\downarrow$ & \textbf{R-Prec (T3)} $\uparrow$ & \textbf{MM-Dist} $\downarrow$ \\
\midrule
MotionGPT~\cite{jiang2023motiongpt}& 0.869 & 0.536 & 3.139 \\
\textbf{ZeroWBC (Ours)} & \textbf{0.756} & \textbf{0.755} & \textbf{3.124} \\
\bottomrule
\end{tabular}%
}
\end{wraptable}
Table~\ref{tab:motion_metrics} compares ZeroWBC with the text-to-motion baseline MotionGPT under the same Nymeria training and testing setting. ZeroWBC achieves better FID, Top-3 R-Precision, and MM-Dist, showing that egocentric visual context improves motion generation beyond language-only conditioning. Additional modality and training-data ablations are reported in the Appendix.

\subsection{Evaluation on General Interactive Motion Tracking}
We evaluate general whole-body motion tracking on two settings: HumanML3D and Generation, as summarized in Table~\ref{tab:motion_tracking_eval}. All motion tracking policies are trained on two NVIDIA RTX 4090 GPUs for approximately 14 days.
\begin{wraptable}[13]{r}{0.65\textwidth}
\vspace{-1.5em}
\centering
\caption{Joint-level motion tracking evaluation. Lower values indicate better performance.}
\vspace{-1.5ex}
\label{tab:motion_tracking_eval}
\scriptsize
\setlength{\tabcolsep}{3pt}
\resizebox{\linewidth}{!}{%
\begin{tabular}{l l c c c c c}
\toprule
Data & Method 
& MPJPE $\downarrow$
& MPJAE $\downarrow$
& MPJVE $\downarrow$
& MPIPE $\downarrow$
& MPIVE $\downarrow$ \\
\midrule

\multirow{6}{*}{\shortstack{HumanML3D}}
& GMT~\cite{chen2025gmt}
& 0.5530 & 0.1046 & 0.4882 & 0.3821 & 0.3216 \\
& SONIC~\cite{luo2025sonic}
& \textbf{0.2526} & \underline{0.0957} & \textbf{0.3906} & \underline{0.2108} & \underline{0.1992} \\
& w/o Curr.& 0.5991 & 0.0953 & 0.6206 & 0.4125 & 0.3501 \\
& w/o A.S. & 0.5808 & 0.1047 & 0.5520 & 0.3987 & 0.3378 \\
& w/o F.C. & 0.4619 & 0.0820 & 0.4912 & 0.3012 & 0.2645 \\
& \textbf{Ours}
& \underline{0.2571} & \textbf{0.0734} & \underline{0.4378} & \textbf{0.1885} & \textbf{0.1753} \\
\midrule

\multirow{6}{*}{\shortstack{Generation}}
& GMT
& 0.6013 & 0.1375 & 0.5112 & 0.4219 & 0.3654 \\
& SONIC
& \textbf{0.3340} & \textbf{0.1058} & \underline{0.4691} & \underline{0.2671} & \underline{0.2318} \\
& w/o Curr. & 0.6637 & 0.1319 & 0.6474 & 0.4789 & 0.4123 \\
& w/o A.S. & 0.6124 & 0.1268 & 0.6293 & 0.4452 & 0.3876 \\
& w/o F.C. & 0.5259 & 0.1137 & 0.4992 & 0.3765 & 0.3296 \\
& \textbf{Ours}
& \underline{0.3653} & \underline{0.1074} & \textbf{0.4680} & \textbf{0.2138} & \textbf{0.2005} \\
\bottomrule
\end{tabular}%
}
\end{wraptable}

HumanML3D follows the definition in Section~\ref{sec:dataset}, while Generation evaluates motions produced by the multimodal motion generation module. We report joint-level tracking metrics in world coordinates: Mean Per-Joint Position Error (MPJPE, m), Mean Per-Joint Angle Error (MPJAE, rad), Mean Per-Joint Velocity Error (MPJVE, rad/s), Mean Per-Interactive-Body-Part Position Error (MPIPE, m), and Mean Per-Interactive-Body-Part Velocity Error (MPIVE, m/s). We compare with GMT~\cite{chen2025gmt} and SONIC~\cite{luo2025sonic}, and include ablations removing current-frame targets (w/o Curr.), adaptive scheduling (w/o A.S.), and future conditioning (w/o F.C.). All methods are evaluated under identical tracking settings. As shown in Table~\ref{tab:motion_tracking_eval}, our method consistently outperforms GMT across both HumanML3D and Generation. Compared with the stronger SONIC baseline, our method remains competitive, achieving the best MPJAE on HumanML3D and the best MPJVE on Generation, while SONIC obtains lower MPJPE on both settings. For both MPIPE and MPIVE, our method achieves the best performance. This trade-off arises because SONIC prioritizes overall joint-level tracking accuracy, whereas our method emphasizes interactive trajectory tracking, leading to superior performance on interaction-related metrics. The ablation results show that all components contribute to robust general motion tracking.


\begin{figure*}[t]
    \centering
    \includegraphics[width=1.0\textwidth]{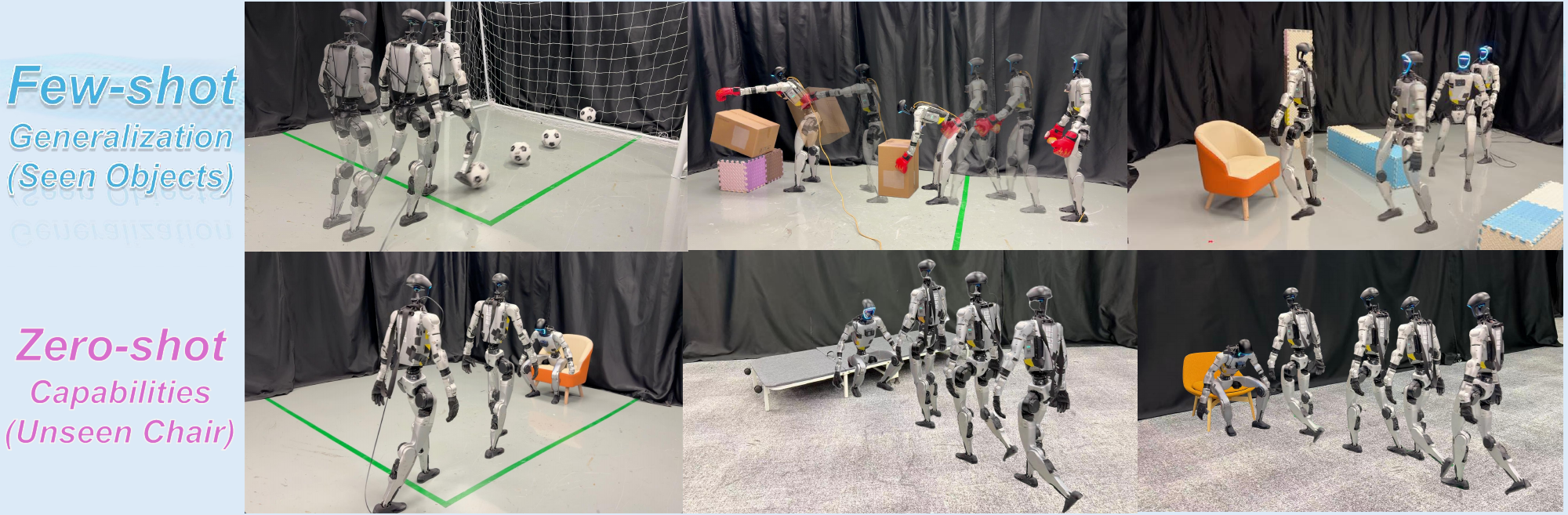}
    \caption{Real-world interaction results on Unitree G1. \textbf{Few-shot generalization}: the robot performs seen task categories under changed object placements, scene layouts, and initial poses. \textbf{Zero-shot capabilities}: the robot performs unseen object-task combinations, including chair navigation and chair sitting, despite the absence of chair-related demonstrations during training.}
    \label{few-shot}
\end{figure*}

We evaluate ZeroWBC on a Unitree G1 humanoid robot across real-world interaction tasks, including navigation, obstacle avoidance, ball kicking, sofa sitting, and box moving. Each episode lasts 10--20 seconds and requires whole-body motion with scene geometry. A GoPro camera mounted on the robot's chest provides the initial egocentric observation, which is sent to a local server running the fine-tuned Qwen-VL model. The generated human motion is decoded, retargeted to the G1 robot, and executed by the tracking policy. Full-system comparisons with other methods are limited because few open-source humanoid interaction methods support whole-body control.

\subsection{Real-World Evaluation}
\begin{wraptable}[6]{r}{0.48\textwidth}
\vspace{-1.2em}
\centering
\caption{Few-shot real-world success rates.}
\vspace{-0.8em}
\label{tab:fewshot_real_world}
\scriptsize
\setlength{\tabcolsep}{3pt}
\resizebox{\linewidth}{!}{%
\begin{tabular}{l c c}
\toprule
\textbf{Task} & \textbf{General Tracker} & \textbf{Interactive Tracker} \\
\midrule
Obstacle Avoidance  & 80\% & \textbf{96\%} \\
Ball Kicking        & 60\% & \textbf{78\%} \\
Sofa Sitting        & 48\% & \textbf{84\%} \\
Box moving          & 16\% & \textbf{64\%} \\
\bottomrule
\end{tabular}%
}
\vspace{-1.0em}
\end{wraptable}
\textbf{Few-shot evaluation.}
Each few-shot task is evaluated over 50 real-world trials, where the task and object categories appear in the training data, but the test environments use different object placements, spatial layouts, and robot initial poses. As shown in Table~\ref{tab:fewshot_real_world}, the interactive tracker achieves high success rates on obstacle avoidance, ball kicking, box moving, and sofa sitting, outperforming the general tracker by better preserving human motion root and key body-part trajectories. The general tracker adopts SONIC as the baseline.

\textbf{Zero-shot evaluation.}
We further evaluate zero-shot generalization on tasks and object categories completely absent from the training data, including unseen chair navigation and chair sitting. Each task is evaluated over 50 trials with varied object poses, robot initial positions, and four different chair types. ZeroWBC achieves success rates of 90.0\% for chair navigation and 76.0\% for chair sitting. These results suggest that Qwen-VL provides useful semantic priors for unseen objects and tasks, while the tracking policy enables feasible whole-body execution.

\section{Conclusion}
\label{sec:conclusion}
We present ZeroWBC, a robot teleoperation-free generation-then-tracking framework that turns human egocentric videos with synchronized whole-body motion and text into executable static humanoid interaction behaviors. By coupling egocentric motion generation with general interactive tracking, ZeroWBC shows that scalable human demonstrations can provide effective supervision for scene-aware whole-body humanoid control in low cost. Experiments in simulation and on Unitree G1 demonstrate natural and diverse behaviors from a single initial image and language instruction, offering a practical path toward learning humanoid interaction without robot teleoperation.

\section{Limitations}
\label{sec:limitations}
ZeroWBC is designed for mostly static scenes. Our experiments show that its open-loop generation-then-tracking strategy works well for static scene-aware tasks, but it cannot replan when objects or humans move after the initial observation. Our closed-loop tests show about 400 ms latency even for short-horizon generation, causing phase lag and jitter. Future work will explore faster generation, model distillation, and closed-loop replanning. Contact-rich tasks remain challenging. In box moving, the robot can often approach and lift the box, but may drop it during transport due to limited hand friction and the lack of force or tactile feedback. Future work will incorporate tactile sensing, force feedback, and contact-aware tracking. Finally, our experiments are limited to Unitree G1 and a small set of tasks; deployment on other humanoids may require embodiment-aware retargeting and tracker adaptation.
\clearpage


\bibliography{example}  

\clearpage
\appendix
\section{Dataset Details}
\label{app:dataset}

We use three datasets in ZeroWBC: Nymeria, HumanML3D, and our self-collected egocentric interaction dataset. All datasets are split into training and testing subsets with an 8:2 ratio at the sequence level. The training split is used for motion tokenizer training, Qwen-VL fine-tuning, and tracking policy training, while the test split is only used for evaluation.

\textbf{Nymeria Dataset.}
Nymeria~\cite{ma2024nymeria} is a large-scale multimodal egocentric dataset containing approximately 300 hours of daily human activities. It provides synchronized egocentric videos, human whole-body motion captured by XSens, and hierarchical language annotations. In our work, we process Nymeria into 5--10 second clips, where each sample consists of an initial egocentric image, a text instruction, and a synchronized SMPL-formatted whole-body motion sequence. After preprocessing, we obtain approximately 140k training samples and 20k testing samples. Nymeria is mainly used for training the VQ-VAE motion tokenizer and for the first-stage fine-tuning of Qwen-VL.

\textbf{HumanML3D Dataset.}
HumanML3D~\cite{Guo_2022_CVPR} contains 14,616 high-quality 3D human motion sequences, corresponding to about 28.59 hours of motion data, paired with precise textual annotations. Since HumanML3D does not contain egocentric images, we use it as text-motion supervision. HumanML3D is used for both training and evaluation in the two stages of ZeroWBC. First, it is co-trained with Nymeria during Qwen-VL fine-tuning to improve text-motion alignment and diversify language-conditioned motion generation. Second, we retarget its human motions to the Unitree G1 robot using GMR~\cite{araujo2025retargeting}, and extract the corresponding global root, hand, and foot trajectories from the original human motions. The retargeted robot motions and these key body-part trajectories are then used to pretrain and evaluate the general interactive tracking policy, providing diverse human whole-body motion priors before task-specific adaptation.

\textbf{Self-collected Dataset.}
To improve spatial grounding and in-domain interactive motion quality, we further collect a human egocentric interaction dataset following the protocol described in Section~\ref{sec:data_collection}. The dataset contains approximately 5 hours of synchronized egocentric video, whole-body motion, and text annotations. It covers four representative static-scene whole-body interaction tasks: box moving, ball kicking, sofa sitting, and obstacle avoidance. Each task contains approximately 300-400 motion sequences, resulting in about 1,500 sequences in total. Each sequence lasts approximately 5--15 seconds, depending on the task. During collection, we vary object positions, scene layouts, backgrounds, and human initial poses to improve robustness to spatial variations. This dataset is mainly used for the second-stage Qwen-VL fine-tuning and task-adaptive tracker fine-tuning, allowing ZeroWBC to better learn object-aware and spatially grounded whole-body interaction behaviors.

\begin{wrapfigure}[17]{r}{0.42\textwidth}
    \vspace{-2.8em}
    \centering
    \includegraphics[width=0.35\textwidth]{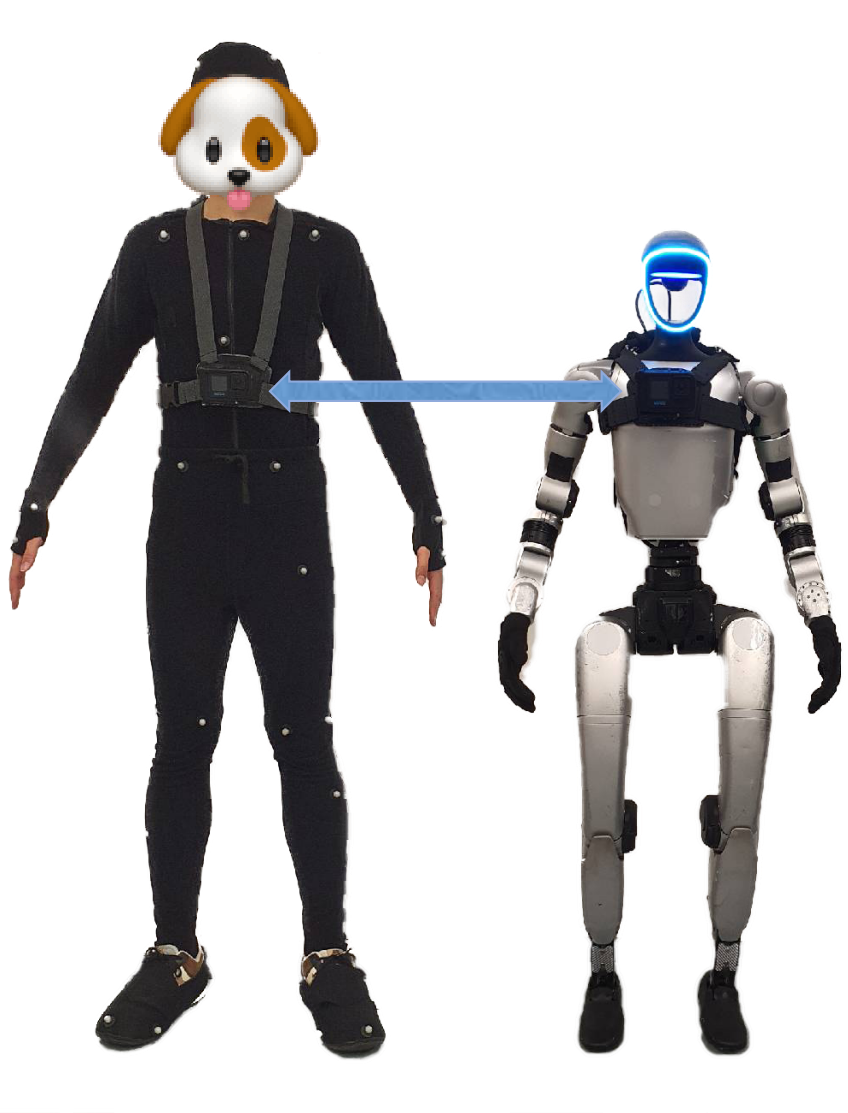}
    \caption{Hardware configuration for perspective alignment. }
    \label{fig:human_humanoid_camera}
\end{wrapfigure}

For the self-collected dataset, whole-body human motion is captured using an optical motion capture system and then converted into the SMPL format. Egocentric images are recorded using a chest-mounted GoPro camera. As shown in Fig.~\ref{fig:human_humanoid_camera}, the camera position is approximately aligned with the camera mounting position on the Unitree G1 robot, so that human demonstrations and robot deployment share similar first-person observations. During data collection, the camera height may vary by approximately $\pm 5$ cm and the pitch angle may differ by roughly $20^\circ$ from the nominal pose. These moderate viewpoint variations improve robustness and reduce overfitting to a single camera configuration. The text annotations are also augmented using GPT-5, which generates semantically equivalent paraphrases for each manually written instruction to increase linguistic diversity while preserving the original task objective and spatial intent.

To ensure the validity of the zero-shot chair evaluation, we remove chair-related samples from all training data. Specifically, we filter out sequences whose text instructions contain chair-related descriptions and remove samples where chairs appear as salient objects in egocentric images. This filtering is applied before both Qwen-VL fine-tuning and interactive tracker training. Therefore, the chair navigation and chair sitting tasks used in real-world zero-shot evaluation are absent from the training data.

We further compare the efficiency of our human egocentric data collection with humanoid whole-body teleoperation. The teleoperation setup uses the SONIC whole-body teleoperation system with PICO 4 Ultra VR control. We use box moving as the comparison task, where the robot moves a box from the ground to a target location and each successful demonstration lasts about 15 seconds. In practice, teleoperation requires two operators for VR control, robot reset, safety monitoring, and task setup. It is also slowed by difficult whole-body control, time-consuming resets, and frequent failures such as failed grasping, box dropping, or loss of balance. As a result, about 8 hours of teleoperation yields only around 100 successful demonstrations. In contrast, one human demonstrator with a motion capture system and a chest-mounted camera can collect about 300 successful egocentric demonstrations in 2 hours for the same task and setting, highlighting the scalability advantage of human egocentric data.

\section{Motion Generation Details and Supplementary Experiments}
\label{app:motion_generation}

We provide additional implementation details of the motion generation module, including the VQ-VAE motion tokenizer, Qwen-VL fine-tuning setup, and supplementary ablation experiments on input modalities and training data.

\begin{wraptable}[17]{r}{0.36\textwidth}
\vspace{-2.0em}
\centering
\caption{VQ-VAE training configuration.}
\label{tab:vqvae_details}
\scriptsize
\renewcommand{\arraystretch}{0.9}
\setlength{\tabcolsep}{2.2pt}
\begin{tabular}{@{}c|l|c@{}}
\toprule
\textbf{Type} & \textbf{Hyper-parameter} & \textbf{Value}\\
\midrule
\multirow{7}{*}{\textbf{Arch}}
& Down-sample stages & 2\\
& Stride per stage & 2\\
& Latent code dim & 512\\
& Code-book size & 2048\\
& Channel width & 512\\
& Residual blocks & 3\\
& Activation & ReLU\\
\midrule
\multirow{5}{*}{\textbf{Loss}}
& Reconstruction $\lambda_r$ & 1.0\\
& Commitment $\lambda_c$ & 0.02\\
& Velocity $\lambda_v$ & 0.10\\
& Root-rotation $\lambda_{rr}$ & 0.5\\
& Root-position $\lambda_p$ & 0.8\\
\midrule
\multirow{7}{*}{\textbf{Optim}}
& Optimizer & AdamW\\
& Learning rate & $2{\times}10^{-4}$\\
& LR decay epochs & 150, 250, 350\\
& Batch size & 256\\
& Grad clipping & 1.0\\
& Mixed precision & FP16\\
& Epochs & 1500\\
\bottomrule
\end{tabular}
\vspace{-1.0em}
\end{wraptable}

\textbf{Motion tokenizer.}
We train a VQ-VAE motion tokenizer to discretize continuous SMPL-formatted human motions into motion tokens for downstream autoregressive generation. Given an input motion clip of length $n_i$, the encoder downsamples it into a latent sequence of length $n_t$, where $n_i=n_t\cdot 2^{n_{\mathrm{down}}}$ and $n_{\mathrm{down}}$ denotes the number of temporal downsampling stages. The latent sequence is quantized by the codebook and then reconstructed by a lightweight decoder to the original motion resolution. Detailed architecture and optimization settings are provided in Table~\ref{tab:vqvae_details}. We use a codebook size of 2048 to balance motion reconstruction fidelity and the prediction difficulty of Qwen-VL.

For each motion sequence $\mathbf{m}$, we denote the SMPL pose as 
$\mathbf{m}_{\mathrm{smpl}}\in\mathbb{R}^{66}$ and the global root translation as 
$\mathbf{m}_{\mathrm{trans}}\in\mathbb{R}^{3}$. The tokenizer is trained with reconstruction, commitment, velocity, root-rotation, and root-position losses:
\begin{align}
\mathcal{L}_{\mathrm{VQ}}
&=
\lambda_r \|\mathbf{m}-\hat{\mathbf{m}}\|_1
+
\lambda_c \|\mathrm{sg}[\mathbf{z}_e]-\mathbf{z}_q\|_2^2
+
\lambda_v \|\Delta\mathbf{m}-\Delta\hat{\mathbf{m}}\|_1
\notag \\
&\quad
+
\lambda_{rr}\|\mathbf{m}_{0:3}-\hat{\mathbf{m}}_{0:3}\|_1
+
\lambda_p\|\mathbf{m}_{\mathrm{trans}}-\hat{\mathbf{m}}_{\mathrm{trans}}\|_1 .
\end{align}
Here, $\hat{\mathbf{m}}$ denotes the reconstructed motion, $\mathbf{z}_e$ is the encoder output, and $\mathbf{z}_q$ is the quantized latent vector selected from the codebook. The operator $\mathrm{sg}[\cdot]$ denotes stop-gradient, and $\Delta(\cdot)$ denotes temporal finite difference. The velocity loss encourages temporal smoothness, while the root-rotation and root-position losses preserve global heading and trajectory information, which is important for spatially grounded interaction. The resulting discrete token sequence is used as the prediction target for Qwen-VL fine-tuning.

\textbf{Qwen-VL Fine-tuning.} We adopt Qwen2.5-VL-3B as the backbone for multimodal motion generation. To formulate motion generation as next-token prediction, we add special motion tokens to the tokenizer vocabulary, including \texttt{<motion\_start>}, \texttt{<motion\_end>}, and discrete codebook tokens such as \texttt{<motion\_token\_0>} and \texttt{<motion\_token\_1>}. Given an initial egocentric image and a language instruction, the model autoregressively predicts a complete future motion token sequence between the start and end tokens. The predicted tokens are then decoded by the VQ-VAE decoder into continuous SMPL motions.

\begin{wrapfigure}[12]{r}{0.42\textwidth}
    \centering
    \vspace{-2.0em}
    \includegraphics[width=1.0\linewidth]{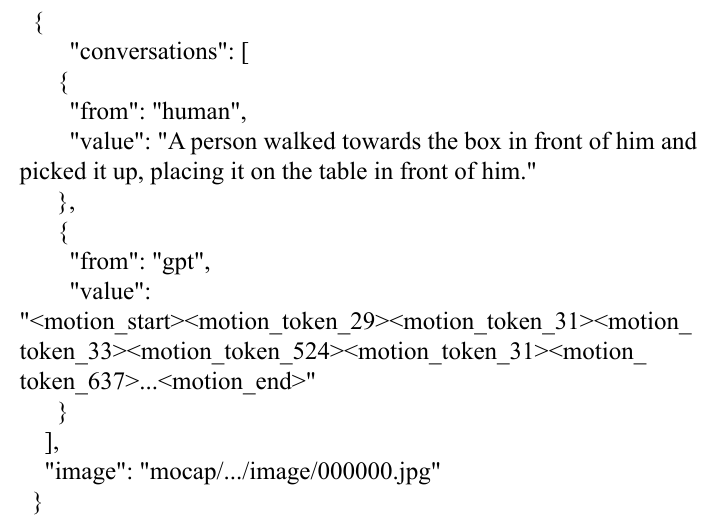}
    \caption{An example fine-tuning sample for multimodal motion generation.}
    \label{fig:finetune_sample}
\end{wrapfigure}

All Qwen-VL fine-tuning experiments are conducted on 32 NVIDIA A100 GPUs. During fine-tuning, we freeze all model parameters except the LM decoder, which is fine-tuned with full parameters. This design preserves the pretrained vision-language alignment ability of Qwen-VL while reducing training cost. The fine-tuning data contains approximately 140k samples from Nymeria, 13k samples from HumanML3D, and 1.5k samples from our self-collected dataset. Nymeria provides large-scale egocentric image-text-motion supervision, HumanML3D improves text-motion alignment, and the self-collected dataset improves spatial grounding for in-domain interactive tasks. Fig.~\ref{fig:finetune_sample} shows a representative fine-tuning sample.

\textbf{Ablations on input modality.} We further evaluate the effect of input modality under the full training setting, where all variants are trained with Nymeria, HumanML3D, and our self-collected dataset. As shown in Table~\ref{tab:modality_ablation}, image-only generation performs the worst because visual observations alone do not fully specify the intended task. Text-only generation achieves better semantic motion quality, but it lacks direct scene grounding and is less effective for spatially dependent interactions. In contrast, image-text conditioning achieves the best performance on both Nymeria and the self-collected test set, showing that language provides task-level semantics while egocentric images provide complementary scene-level spatial context. Since existing multimodal motion generation baselines rarely support human motion generation conditioned on both first-person visual observations and language instructions, direct comparison is limited. We therefore evaluate the contribution of each modality through controlled image-only and text-only ablations.
\begin{table}[t]
    \centering
    \caption{Ablations on input modality for motion generation. All variants are trained with Nymeria, HumanML3D, and the self-collected dataset.}
    \label{tab:modality_ablation}
    \footnotesize
    \setlength{\tabcolsep}{6.0pt}
    \renewcommand{\arraystretch}{1.08}
    \begin{tabular}{c c c c c}
        \toprule
        Test Set & Modality & FID $\downarrow$ & R-Prec (T3) $\uparrow$ & MM-Dist $\downarrow$ \\
        \midrule
        \multirow{3}{*}{Nymeria}
        & Image only  & 0.925 & 0.534 & 3.854 \\
        & Text only   & 0.518 & 0.685 & 2.864 \\
        & Image + Text & \textbf{0.298} & \textbf{0.847} & \textbf{2.286} \\
        \midrule
        \multirow{3}{*}{Self-collected}
        & Image only  & 0.563 & 0.642 & 3.015 \\
        & Text only   & 0.495 & 0.715 & 2.954 \\
        & Image + Text & \textbf{0.245} & \textbf{0.892} & \textbf{2.456} \\
        \bottomrule
    \end{tabular}
    \vspace{-0.6em}
\end{table}

\section{General Interactive Motion Tracking Details and Experiments}
\label{app:tracking_details}

In this section, we provide the specific configuration for the General Motion Tracking policy, including the mathematical formulation of the reward functions and the curriculum learning strategy based on motion difficulty grading.

\textbf{Reward Function Configuration.} The total reward $r_t$ is a weighted sum of task rewards (tracking precision) and regularization rewards (safety and smoothness). The formulation involves standard geometric computations, where $\mathbf{p}$ and $\mathbf{q}$ denote position and orientation (quaternion), $\mathbf{v}$ and $\boldsymbol{\omega}$ denote linear and angular velocities. The hat notation (e.g., $\hat{\mathbf{p}}$) indicates the robot's actual state, while variables without hats represent the reference motion. The symbol $\ominus$ represents the quaternion error magnitude. 

Task rewards penalize deviations from the reference motion using an exponential kernel $\exp(-d^2/\sigma^2)$, where $d$ is the error metric and $\sigma$ controls the sensitivity. The specific terms and parameters used in our best-performing model are detailed in Table \ref{tab:task_rewards}.

\begin{table}[t]
    \centering
    \caption{Task Reward Terms. All terms are defined based on the unified tracking targets $\boldsymbol{g}_{\text{inter}}$ (interaction trajectories) and $\boldsymbol{m}_{\text{ref}}$ (reference motion). $N$ denotes the number of tracked body parts.}
    \label{tab:task_rewards}
    
    \renewcommand{\arraystretch}{1.5} 
    \resizebox{\columnwidth}{!}{%
    \begin{tabular}{l c c c}
        \toprule
        \textbf{Reward Term} & \textbf{Expression} & \textbf{Weight} & \textbf{Sens. ($\sigma$)} \\
        \midrule
        Wrist Interaction Pos. & $\exp\left(-\frac{\|\mathbf{p}_{\text{wrist}} - \hat{\mathbf{p}}_{\text{wrist}}\|^2}{\sigma^2}\right)$ & 1.0 & 0.3 \\
        Wrist Interaction Vel. & $\exp\left(-\frac{\|\mathbf{\dot{p}}_{\text{wrist}} - \hat{\mathbf{\dot{p}}}_{\text{wrist}}\|^2}{\sigma^2}\right)$ & 1.0 & 0.2 \\
        Foot Interaction Pos. & $\exp\left(-\frac{\|\mathbf{p}_{\text{foot}} - \hat{\mathbf{p}}_{\text{foot}}\|^2}{\sigma^2}\right)$ & 1.0 & 0.3 \\
        Foot Interaction Vel. & $\exp\left(-\frac{\|\mathbf{\dot{p}}_{\text{foot}} - \hat{\mathbf{\dot{p}}}_{\text{foot}}\|^2}{\sigma^2}\right)$ & 1.0 & 0.2 \\
        Root Position & $\exp\left(-\frac{\|\mathbf{p}_{\text{root}} - \hat{\mathbf{p}}_{\text{root}}\|^2}{\sigma^2}\right)$ & 0.8 & 0.2 \\
        Root Orientation Velocity & $\exp\left(-\frac{\|w_{\text{root}} -\hat{w}_{\text{root}}\|^2}{\sigma^2}\right)$ & 0.8 & 0.4 \\
        Reference Joint Pos. & $\exp\left(-\frac{\frac{1}{N}\sum_{i}\|\mathbf{q}_{\text{ref},i} - \hat{\mathbf{q}}_{\text{ref},i}\|^2}{\sigma^2}\right)$ & 0.1 & 0.3 \\
        Reference Joint Ang. Vel. & $\exp\left(-\frac{\frac{1}{N}\sum_{i}\|\boldsymbol{\dot q}_{\text{ref},i} - \hat{\boldsymbol{\dot q}}_{\text{ref},i}\|^2}{\sigma^2}\right)$ & 0.4 & 3.14 \\
        \bottomrule
    \end{tabular}%
    }
\end{table}

Regularization terms prevent unphysical behaviors and ensure hardware safety. These are implemented as penalties (negative rewards) or constraints, as shown in Table \ref{tab:reg_rewards}. Notably, the \textit{Undesired Contacts} term penalizes contact forces $F_b > 1.0$N on all body parts except the feet (ankles) and hands (wrists), allowing for locomotion and potential hand-ground interactions (e.g., crawling).

\begin{wraptable}[8]{r}{0.50\textwidth}
\centering
\vspace{-1.0em}
\caption{Regularization and penalty terms.}
\label{tab:reg_rewards}
\scriptsize
\setlength{\tabcolsep}{3pt}
\renewcommand{\arraystretch}{1.2}
\resizebox{\linewidth}{!}{%
\begin{tabular}{l l c}
\toprule
\textbf{Term} & \textbf{Expression / Logic} & \textbf{Weight} \\
\midrule
Action Rate L2 
& $-\|a_t - a_{t-1}\|_2^2$ 
& -0.1 \\
Joint Limit 
& $-\sum_{j} \mathbf{1}_{out}(q_j)$ 
& -10.0 \\
Undesired Contacts 
& $-\sum_{b \notin \mathcal{B}_{allow}} \mathbf{1}_{F_b > 1.0}$ 
& -0.1 \\
\bottomrule
\end{tabular}%
}
\end{wraptable}

\textbf{Motion Difficulty Curriculum.} We categorize motions into difficulty levels from 1 to 10 based on their motion intensity, coordination complexity, and stability requirements.
Lower difficulty levels correspond to static or simple locomotion motions, while higher levels involve dynamic, low-stability, or acrobatic actions.
Table~\ref{tab:difficulty_curriculum} summarizes the grading criteria and representative motions.

\begin{wraptable}[13]{r}{0.50\textwidth}
\vspace{-3.0em}
\centering
\caption{Actor observation space.}
\label{tab:obs_space}
\scriptsize
\setlength{\tabcolsep}{3pt}
\renewcommand{\arraystretch}{1.1}
\resizebox{\linewidth}{!}{%
\begin{tabular}{l l c}
\toprule
\textbf{Component} & \textbf{Description} & \textbf{Dim} \\
\midrule
\multicolumn{3}{l}{\emph{Motion Command}} \\
current\_target & Current target interactive motion & 98 \\
short\_future\_targets & Short-horizon targets (2 frames) & 196 \\
long\_future\_targets & Long-horizon targets (5 frames) & 490 \\
\midrule
command (total) & Multi-scale motion target & 784 \\
\midrule
\multicolumn{3}{l}{\emph{Proprioceptive Observations}} \\
motion\_anchor\_ori\_b & Root orientation in body frame & 6 \\
base\_ang\_vel & Base angular velocity & 3 \\
joint\_pos & Joint positions & 29 \\
joint\_vel & Joint velocities & 29 \\
prev\_actions & Previous action & 29 \\
\midrule
Total & Actor observation dimension & 880 \\
\bottomrule
\end{tabular}%
}
\vspace{-1.0em}
\end{wraptable}

During training, the tracking policy is exposed to motions with gradually increasing difficulty.
The policy first learns to track low-difficulty motions and is progressively trained on higher-difficulty motions as tracking performance improves.
This curriculum strategy encourages stable learning and improves robustness when tracking complex motions.

The three difficulty levels illustrated in Fig.~\ref{pipeline} are used only for visualization.
Specifically, Fig.~\ref{pipeline} Level~1 corresponds to difficulty ratings 1--4,
Level~2 corresponds to ratings 5--7,
and Level~3 corresponds to ratings 8--10.
The actual training curriculum operates on the full 1--10 difficulty scale.

\begin{table}[t]
    \centering
    \caption{Motion Difficulty Grading System.}
    \label{tab:difficulty_curriculum}
    \renewcommand{\arraystretch}{1.2} 
    \small 
    \begin{tabularx}{\columnwidth}{c >{\raggedright\arraybackslash}X >{\raggedright\arraybackslash}X}
        \toprule
        \textbf{Rating} & \textbf{Representative Actions} & \textbf{Interpretation} \\
        \midrule
        1 & Stand, raise arms, wave hands & Mostly static or upper-body actions with low motion intensity. \\
        \addlinespace 
        2--4 & Walk, step, turn & Basic locomotion with stable rhythm and simple motion patterns. \\
        \addlinespace
        5--7 & Walk, run, jump, walk backwards & Combined locomotion and dynamic actions requiring higher coordination. \\
        \addlinespace
        8 & Crawl, kneel, crawl on knees & Low-posture ground movements with increased body control demands. \\
        \addlinespace
        9--10 & Cartwheel, backflip, jump, dance & High-skill or acrobatic actions with the highest complexity and risk. \\
        \bottomrule
    \end{tabularx}
\end{table}

\begin{table}[H]
\centering
\caption{Joint-level motion tracking evaluation. Lower values indicate better performance.}
\label{tab:motion_tracking_eval}
\tiny
\renewcommand{\arraystretch}{0.92}
\setlength{\tabcolsep}{3pt}
\resizebox{\linewidth}{!}{%
\begin{tabular}{l l c c c c c}
\toprule
Data & Method 
& MPJPE $\downarrow$
& MPJAE $\downarrow$
& MPJVE $\downarrow$
& MPIPE $\downarrow$
& MPIVE $\downarrow$ \\
\midrule

\multirow{7}{*}{\shortstack{HumanML3D}}
& GMT~\cite{chen2025gmt}
& 0.5530 & 0.1046 & 0.4882 & 0.3821 & 0.3216 \\
& SONIC~\cite{luo2025sonic}
& \underline{0.2526} & 0.0957 & \textbf{0.3906} & \underline{0.2108} & 0.1992 \\
& w/o Curr.& 0.5991 & 0.0953 & 0.6206 & 0.4125 & 0.3501 \\
& w/o A.S. & 0.5808 & 0.1047 & 0.5520 & 0.3987 & 0.3378 \\
& w/o F.C. & 0.4619 & 0.0820 & 0.4912 & 0.3012 & 0.2645 \\
& w/o Interact & \textbf{0.2339} & \textbf{0.0691} & \underline{0.4212} & 0.2165 & \underline{0.1894} \\
& \textbf{Ours}
& 0.2571 & \underline{0.0734} & 0.4378 & \textbf{0.1885} & \textbf{0.1753} \\
\midrule

\multirow{7}{*}{\shortstack{Self-Collected}}
& GMT
& 0.5478 & 0.0989 & 0.4725 & 0.3568 & 0.3007 \\
& SONIC
& 0.2493 & \textbf{0.0716} & \underline{0.4021} & \underline{0.2057} & \underline{0.1904} \\
& w/o Curr.& 0.5584 & 0.0917 & 0.5918 & 0.3896 & 0.3312 \\
& w/o A.S. & 0.5415 & 0.0995 & 0.5237 & 0.3741 & 0.3189 \\
& w/o F.C. & 0.4356 & 0.0801 & 0.4674 & 0.2865 & 0.2498 \\
& w/o Interact & \textbf{0.2196} & \underline{0.0737} & \textbf{0.3917} & 0.2175 & 0.1962 \\
& \textbf{Ours}
& \underline{0.2381} & 0.0752 & 0.4196 & \textbf{0.1819} & \textbf{0.1684} \\
\midrule

\multirow{7}{*}{\shortstack{Generation}}
& GMT
& 0.6013 & 0.1375 & 0.5112 & 0.4219 & 0.3654 \\
& SONIC
& \textbf{0.3340} & \underline{0.1058} & 0.4691 & \underline{0.2671} & \underline{0.2318} \\
& w/o Curr. & 0.6637 & 0.1319 & 0.6474 & 0.4789 & 0.4123 \\
& w/o A.S. & 0.6124 & 0.1268 & 0.6293 & 0.4452 & 0.3876 \\
& w/o F.C. & 0.5259 & 0.1137 & 0.4992 & 0.3765 & 0.3296 \\
& w/o Interact & \underline{0.3585} & \textbf{0.0924} & \textbf{0.4537} & 0.2834 & 0.2481 \\
& \textbf{Ours}
& 0.3653 & 0.1074 & \underline{0.4680} & \textbf{0.2138} & \textbf{0.2005} \\
\bottomrule
\end{tabular}%
}
\end{table}

\textbf{Actor Observation Space.}
The actor uses a compact observation space available at deployment time. As shown in Table~\ref{tab:obs_space}, the observation consists of a multi-scale motion command and proprioceptive states. The motion command includes the interactive trajectories of end-effectors and root part, current target joint pose, short-horizon future targets, and long-horizon future targets. The long-horizon targets are further encoded by a lightweight temporal encoder before being combined with proprioceptive observations, enabling the policy to capture global motion trends while preserving near-term tracking accuracy.

\textbf{Full Evaluation on Motion tracking.}
Table~\ref{tab:motion_tracking_eval} reports the full motion tracking results on HumanML3D, self-collected motions, and generated motions. Compared with GMT, our method achieves consistently better tracking performance across all three evaluation settings. Compared with the stronger SONIC baseline, our method is not always the best on standard joint-level metrics such as MPJPE, because our tracker does not solely optimize joint-space imitation. Instead, it explicitly prioritizes task-critical root, wrist, and foot trajectories. This design leads to clear improvements on the interaction-related metrics MPIPE and MPIVE, where our method achieves the best results across all datasets. The comparison with \textit{w/o Interact} further shows this trade-off: removing interaction-oriented targets can slightly improve some joint-level metrics, but substantially degrades key body-part tracking accuracy. These results support our main claim that accurate interaction requires preserving global root and key body-part trajectories, rather than only tracking retargeted joint motions.

\section{Real-world Evaluation Details}
\label{app:real_world}
We provide additional details of the real-world deployment and evaluation protocol on the Unitree G1 humanoid robot. In our system, the Unitree G1 client captures the initial egocentric image from the onboard chest-mounted camera and sends it to a local server equipped with an NVIDIA RTX 4090 GPU. The server runs the fine-tuned Qwen-VL model and generates a complete future human motion sequence from the initial image and language instruction. The generated motion is then decoded, retargeted to the Unitree G1, and sent back to the robot for execution. The interactive tracking policy is deployed directly on the G1 onboard computer, which tracks the retargeted robot motion together with the root and key body-part trajectories. Therefore, high-level motion generation is performed on the external server, while low-level motion tracking and control are executed locally on the robot.

\begin{figure}[h]
    \centering
   \includegraphics[width=1.0\linewidth]{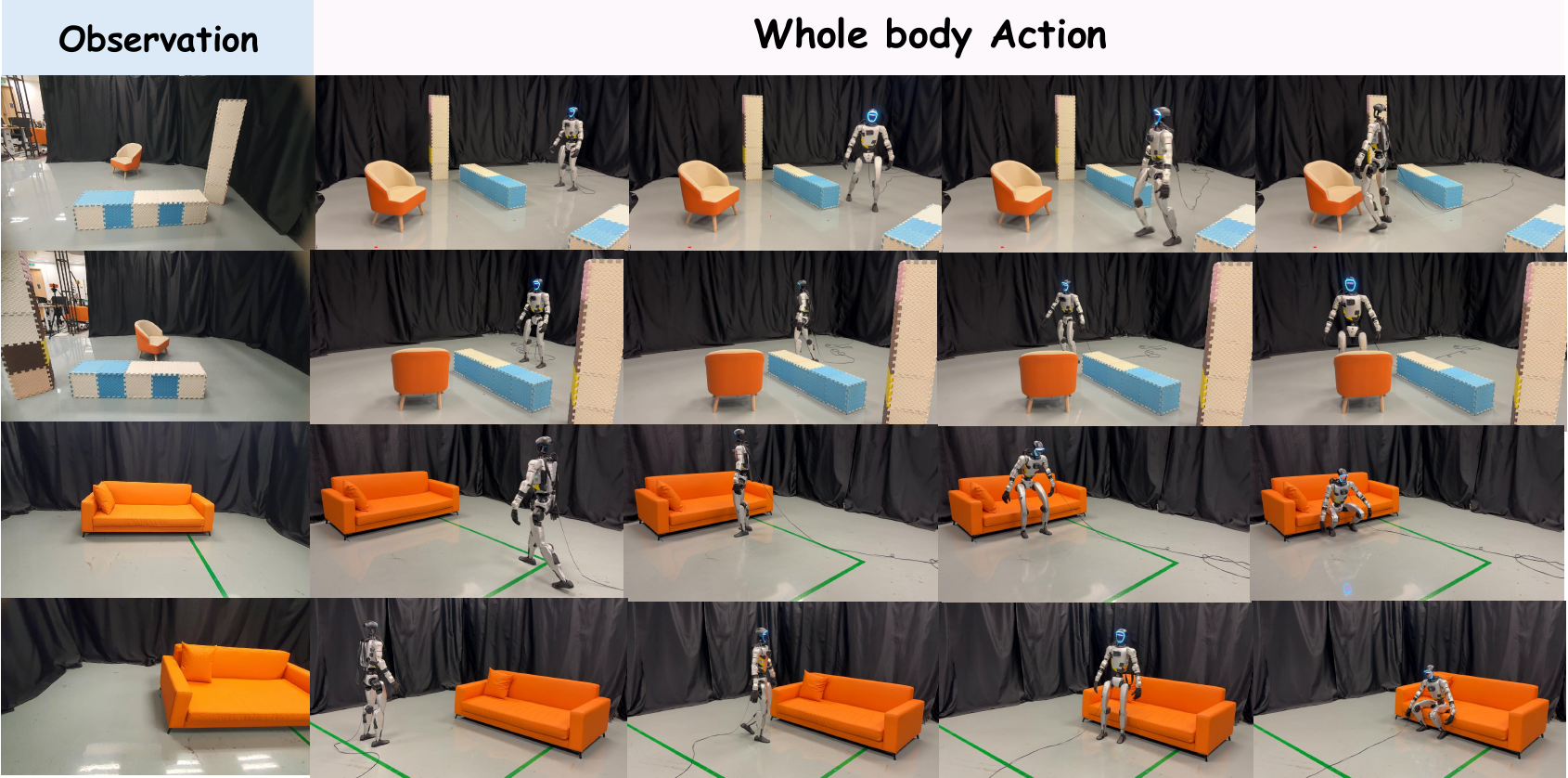}
    \caption{Real-world evaluation under varied initial observations and spatial configurations.  Each row shows the robot execution from a different initial egocentric observation, robot starting pose, and object placement. }
    \label{fig:obs}
\end{figure}
For all real-world evaluations, a trial is counted as successful only if the humanoid completes the entire task from the initial position to the target outcome without falling, unsafe collision, or manual intervention. The success or failure of each trial is determined by human inspection based on whether the intended task is completed. In both few-shot and zero-shot evaluations, the robot's initial position and object placement are randomly initialized for each trial, as illustrated in Fig.~\ref{fig:obs}. The only constraint is that the target object or relevant scene structure must appear in the robot's initial egocentric view. This protocol prevents the robot from simply replaying a fixed trajectory and evaluates whether the generated motion can adapt to different spatial configurations.

Direct comparison with existing methods is limited because there are currently few open-source humanoid systems that support whole-body interaction. Most prior humanoid control systems focus on locomotion, teleoperation, third-person imitation, or task-specific simulation policies, making them not directly comparable to our setting. We therefore conduct a real-world input-modality ablation to evaluate the necessity of multimodal conditioning. With text-only input, the model can infer the task semantics but cannot perceive the spatial layout of the scene, often producing motions that are semantically plausible but spatially misaligned with the object. With image-only input, the model observes the scene but lacks the task instruction, making the intended behavior ambiguous. In both cases, the real-world success rate is nearly zero. These results indicate that both egocentric visual observations and language instructions are necessary for reliable scene-aware whole-body interaction.

\section{Additional analysis}
\label{Additional analysis}
\textbf{Feasibility Analysis of Real-time Control.} We further investigate the feasibility of real-time closed-loop control with visual feedback. By limiting each inference step to a short horizon of 4 motion tokens, the Qwen-VL inference time can be reduced to about 300 ms on an NVIDIA RTX 4090 GPU. However, the end-to-end latency approaches 400 ms after including image transmission, token-to-motion decoding, and retargeting. Such latency is too high for high-frequency feedback control and can cause phase lag, delayed reactions, and control jitter during real-world execution. Therefore, under the current latency of large vision-language models, we adopt an open-loop generation-then-tracking strategy for mostly static scenes. The complete motion sequence is generated once from the initial observation and then executed by the low-level tracker, which decouples high-level semantic planning from real-time control. As the high success rate shown in Table \ref{tab:fewshot_real_world}, this design enables smoother and more stable long-horizon execution in static interaction tasks. Future work will explore faster motion generation, model distillation, and closed-loop replanning for dynamic environments.

\begin{wrapfigure}[22]{r}{0.50\textwidth}
    \vspace{-1.0em}
    \centering
    \includegraphics[width=0.50\textwidth]{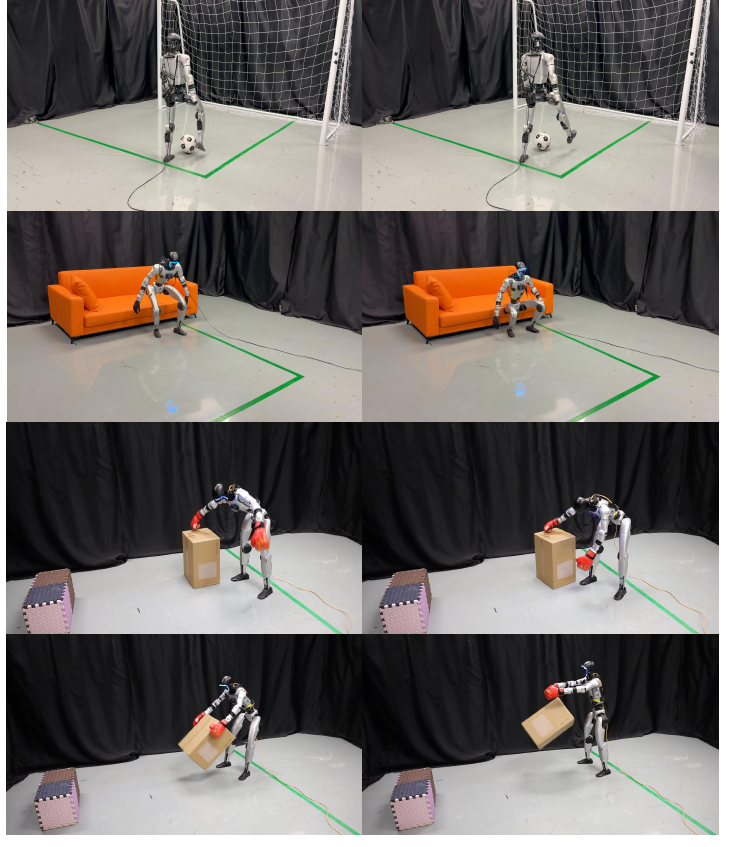}
    \caption{Failure Case Analysis for Box Moving, Sofa Sitting, and Ball Kicking. }
    \label{fig:fail_case}
\end{wrapfigure}

\textbf{Failure Case Analysis.} Most failures arise from contact limitations or small spatial misalignments during execution. For example, in the box-moving task, the robot often approaches and lifts the box correctly, indicating that the generated motion captures the intended visual and semantic goal. However, failures can occur when the box slips during transport due to limited hand friction and the lack of force or tactile feedback. As shown in Fig.~\ref{fig:fail_case}, in other tasks, such as ball kicking or sofa sitting, failures are usually caused by small spatial errors between the generated body trajectory and the target object, e.g., slight foot-ball misalignment or off-center sitting. Even in these cases, the robot typically executes smooth and natural whole-body motions, but the final task outcome may fail due to the precision required by contact-rich interaction. These observations suggest that ZeroWBC can generate semantically meaningful motions, while future improvements should focus on contact-aware tracking, tactile feedback, and closed-loop spatial correction.

\textbf{Zero-shot Generalization Tasks.} It is important to highlight that the task of ``sitting on the chair'' presents significantly higher complexity than ``sitting on the sofa.'' While a sofa offers a broad, forgiving target area, a single-seat chair imposes strict geometric constraints: its width is nearly identical to that of the humanoid's torso. Consequently, the margin for error is minimal, necessitating highly precise maneuvering and spatial alignment. Despite this absence and the lack of such fine-grained examples, the robot successfully generalized the ``sitting'' affordance to different chair styles. This demonstrates the powerful capability of the pre-trained Vision-Language Model, which can adapt learned behaviors (sitting) to unseen objects with far stricter spatial tolerances.

\end{document}